\documentclass[letterpaper]{article} 
\usepackage{aaai23}  
\usepackage{times}  
\usepackage{helvet}  
\usepackage{courier}  
\usepackage[hyphens]{url}  
\usepackage{graphicx} 
\usepackage{color}
\usepackage{natbib}  
\usepackage{caption} 
\usepackage{newfloat}
\usepackage{listings}
\DeclareCaptionStyle{ruled}{labelfont=normalfont,labelsep=colon,strut=off} 
\usepackage{xcolor}

\title{Augmenting Flight Training with AI to Efficiently Train Pilots}
\author {
     Michael Guevarra\textsuperscript{\rm 1},
     Srijita Das\textsuperscript{\rm 2,3},
     Christabel Wayllace\textsuperscript{\rm 2,3},
     Carrie Demmans Epp\textsuperscript{\rm 2},\\
     Matthew E.~Taylor\textsuperscript{\rm 2,3},
     Alan Tay\textsuperscript{\rm 1}
 }
 \affiliations {
     \textsuperscript{\rm 1}  Delphi Technology Corp:  \texttt{guevarrm@myumanitoba.ca}, \texttt{alantay@delphitechcorp.com}\\
     \textsuperscript{\rm 2} University of Alberta: \{\texttt{srijita1, wayllace, demmanse, matthew.e.taylor}\}@\texttt{ualberta.ca}\\
     \textsuperscript{\rm 3} Alberta Machine Intelligence Institute (Amii)
}

\begin{document}

\maketitle

\begin{abstract}
We propose an AI-based pilot trainer to help students learn how to fly aircraft. First, an AI agent uses behavioral cloning to learn flying maneuvers from qualified flight instructors. Later, the system uses the agent's decisions to detect errors made by students and provide feedback to help students correct their errors. This paper presents an instantiation of the pilot trainer. We focus on teaching straight and level flying maneuvers by automatically providing formative feedback to the human student. 
\end{abstract}


There is a critical shortage of commercial pilots worldwide: according to~\citet{Wyman:22}, there will be a global gap of 34,000 pilots by 2025. Part of the problem is that pilots qualified to conduct such training are in very high demand and in short supply. Currently, human instructors guide trainees using flight simulator exercises. We posit that training an AI-enabled system to provide instruction for some tasks is a viable approach to reducing instructor workload while allowing them to interact with more students. This could increase the number of students per pilot trainer, improving the throughput of training pilots and therefore increase the supply of trained pilots.

In recent human-in-the-loop research, AI agents use advice from humans in different forms to speed up learning~\cite{bignold21,cui:21,ChristianoLBMLA17,da2020uncertainty}. Specifically, imitation learning allows an agent to learn to mimic a human’s behavior. Further, AI has been used for airplane flying~\cite{morales2004learning, sandstrom2022fighter} as well as inside intelligent tutoring systems for multiple tasks ranging from student skill development~\cite{georgila2019using} to improving teaching strategies~\cite{wang2018reinforcement}.  This paper presents a system where an agent mimics a qualified pilot and assists students in a pilot training program. Specifically, we focus on the straight and level flight task as a proof of concept. A trained agent identifies mistakes or sub-optimal maneuvers of trainee pilots inside a flight simulator and suggests corrective actions. To the best of our knowledge, this is a first attempt to use an AI tutor to train human pilots for flight. 

\section{System Architecture}
\begin{figure}[t]
	\begin{minipage}{1\linewidth}
		\centering \small
		\includegraphics[width=\textwidth]{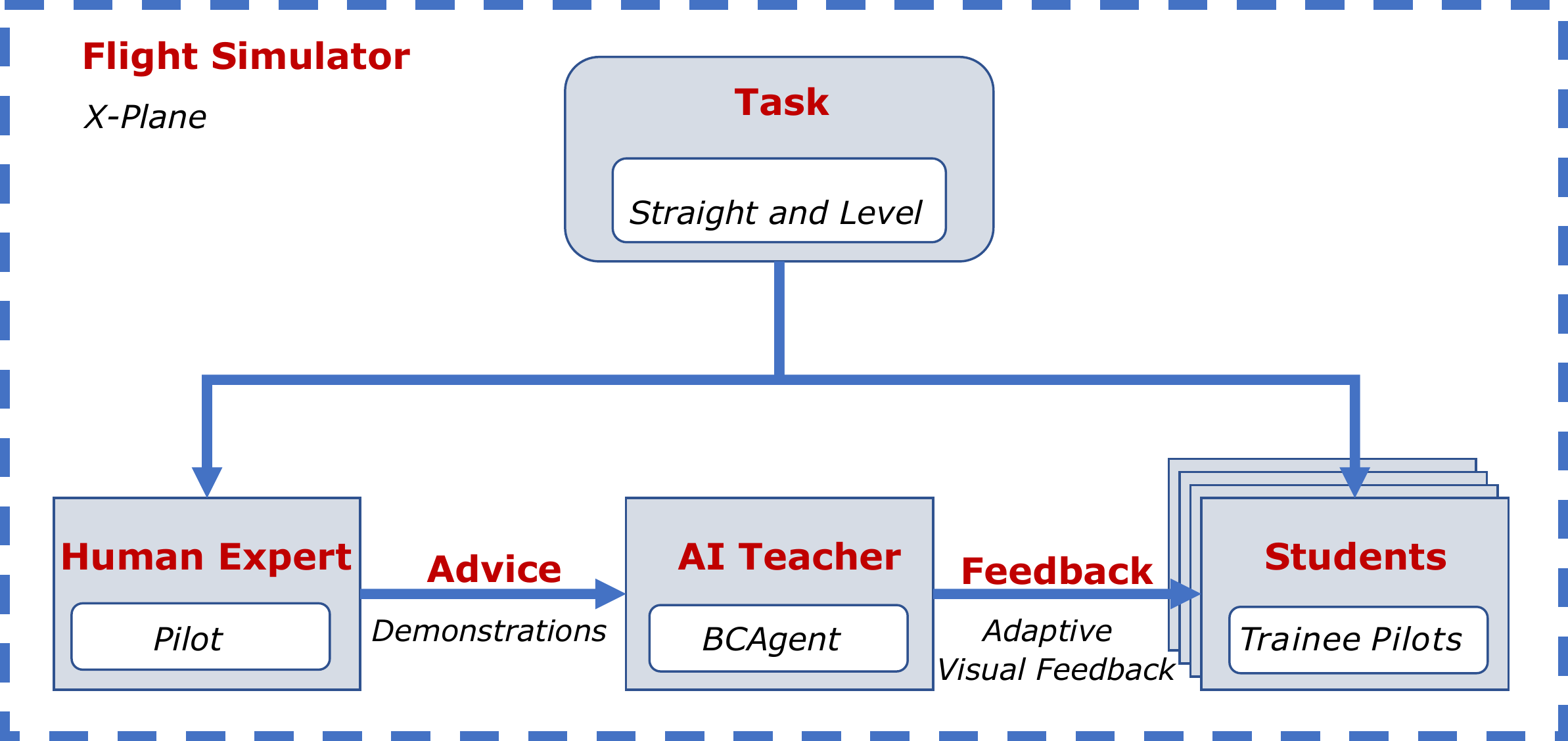} 
		\caption{System Architecture} 
		\label{fig:system}
	\end{minipage}
\end{figure}
 
Our proposed intelligent tutoring system is shown in Figure~\ref{fig:system}. It includes four general components: (1) Task, (2) Human expert, (3) AI teacher, and (4) Students. The task is sampled from a curriculum of flying maneuvers useful for learning to fly. Both the pilot and students perform tasks using the flight simulator X-Plane.\footnote{https://developer.x-plane.com/article/airport-data-apt-dat-file-format-specification/} We use the fundamental flight maneuver ``straight and level"\footnote{\label{note1}Straight and level flight is a flight in which a constant heading and altitude are maintained. It is an essential flight maneuver used to form correct habits. All other flight maneuvers derive from straight and level.} as the target task. There is a human expert (pilot) who is adept in the task and can provide advice to train the AI teacher. After the AI teacher is trained to mimic the human teacher, the AI teacher is used to guide students by providing different types of feedback based on their performance on the target task. 

\subsection{Modeling the Pilot: Agent Training}

We trained a decision-making agent to learn from pilot demonstrated trajectories. Experts demonstrated the``straight and level task" inside the flight simulator for $12.5$ minutes. They operate under visual flight rules, i.e, clear weather to fly towards the target. The target direction is changed between every trial demonstrated by the pilot to account for diversity in trajectory collection. The final dataset consists of $25$ trials of $30$ seconds each with a randomized goal between $\pm 30$ degrees from the starting heading.

\begin{figure}[t]
	\begin{minipage}{1\linewidth}
		\centering \small
		\includegraphics[width=\textwidth]{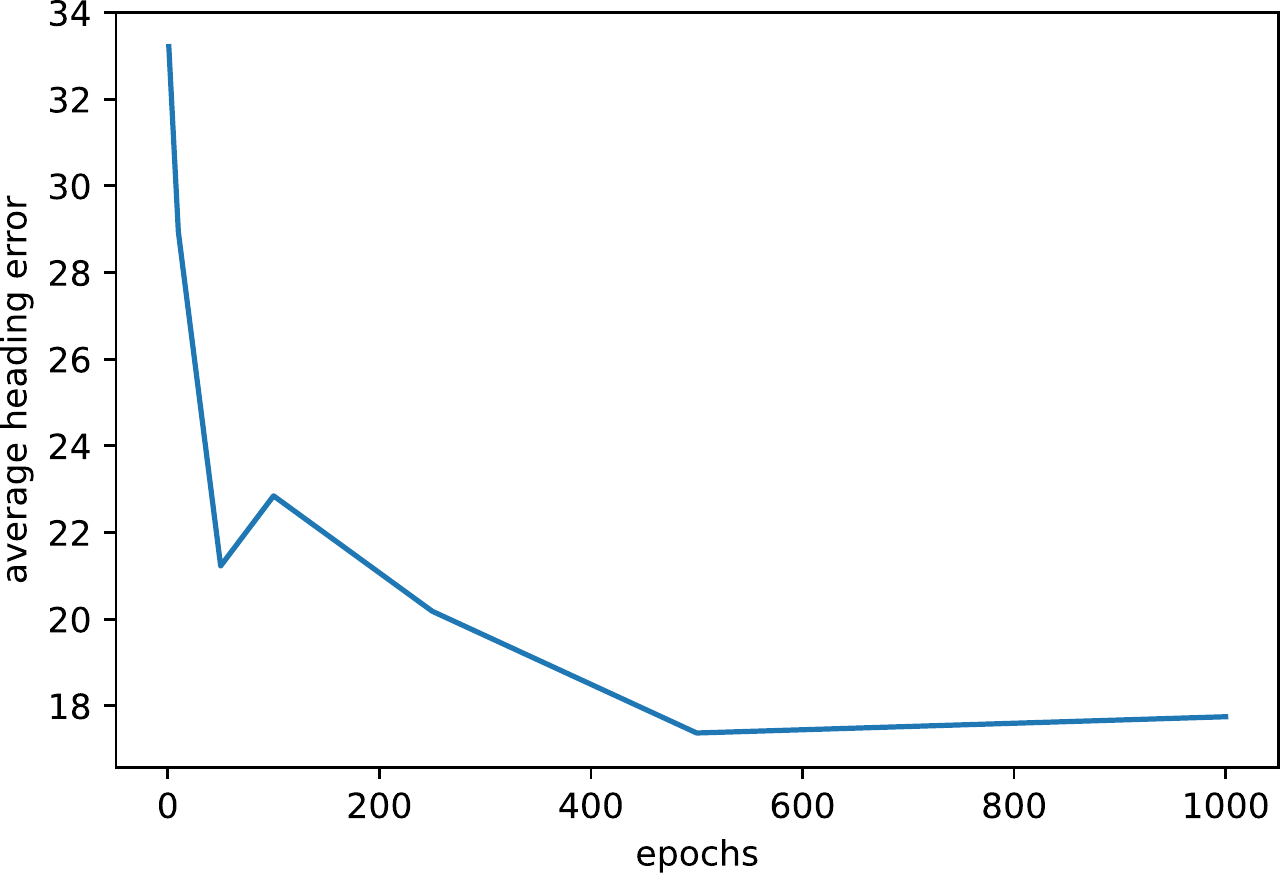} 
		\caption{Average training time error of the behavior cloning teacher}
		\label{fig:evaluate}
	\end{minipage}
\end{figure}

After collecting demonstrations, we trained a behavior cloning (BC) agent~\cite{pomerleau1988alvinn} to mimic the pilot's policy. It is the simplest imitation learning technique where supervised learning is used to mimic the actions of an expert. The BC loss is defined as 
\begin{equation}
   L(\theta)=\sum_{i=1}^{N}||\pi_{\theta}(s_i)-\pi_e(s_i)||^2
   \label{bc_loss}
\end{equation}
where $\pi_{\theta}(s_i)$ is the current policy, $\pi_e(s_i)$ is the expert policy, $\theta$ is the training model parameter, and $N$ refers to the number of state-action pairs in the training set. We used stable baselines~\cite{stable-baselines3} to train the agent from demonstrations. The agent predicts the pitch and roll of the aircraft yoke. We evaluate the agent by measuring the average heading error over $10$ evaluation trials with randomized heading as seen in Figure~\ref{fig:evaluate} and terminate when the error stops improving.


\subsection{Deploying the Teacher and Guiding Students}

We deploy an agent when it makes similar decisions to the expert in previously unseen trials. Mimicking an expert pilot's behavior is not enough to teach students. A teacher should be able to detect students' mistakes and provide feedback to correct them. Therefore, we: (1)~Recorded students' poorly performed flights. (2)~Asked the pilots to prepare annotated critiques on errors made. (3)~Identified two main types of errors. (4)~Used simple distance metrics to decide whether the agent agreed or not with the student's decisions.

The identified student errors are due to (1) Not keeping altitude and airspeed constant and (2) Overshooting the target. To identify them, we compared how the agent and the student controlled the pitch and roll.  

Let $p_a(t), r_a(t), p_s(t),$ and~$r_s(t)$ represent the pitch and roll produced by the agent and the student at a given time~$t$. Then, for the first type of error:
\begin{equation}
  |p_a(t) - p_s(t)| \ge D_1  
  \label{eq_error1}
\end{equation}
where $D_1 \ge 0$ is a user-defined threshold. That is, the pitch difference is larger than the user-defined threshold. 

For the second type of error:
\begin{equation}
    |r_a(t) - r_s(t)| \ge D_2
    \label{eq_error2}
\end{equation}
where $D_2 \ge 0$ is a user-defined threshold. That is, the roll difference is larger than the user-defined threshold.

The agent uses Eqs.~\ref{eq_error1} and~\ref{eq_error2} to determine when to provide feedback. With respect to the type of feedback, this paper presents an instance of \emph{informative tutoring}, a type of formative feedback that presents verification feedback, error flagging, and strategic hints on how to proceed~\citep{shute:08}.
\begin{figure}[t]
	\begin{minipage}{1\linewidth}
		\centering \small
		\includegraphics[width=\textwidth]{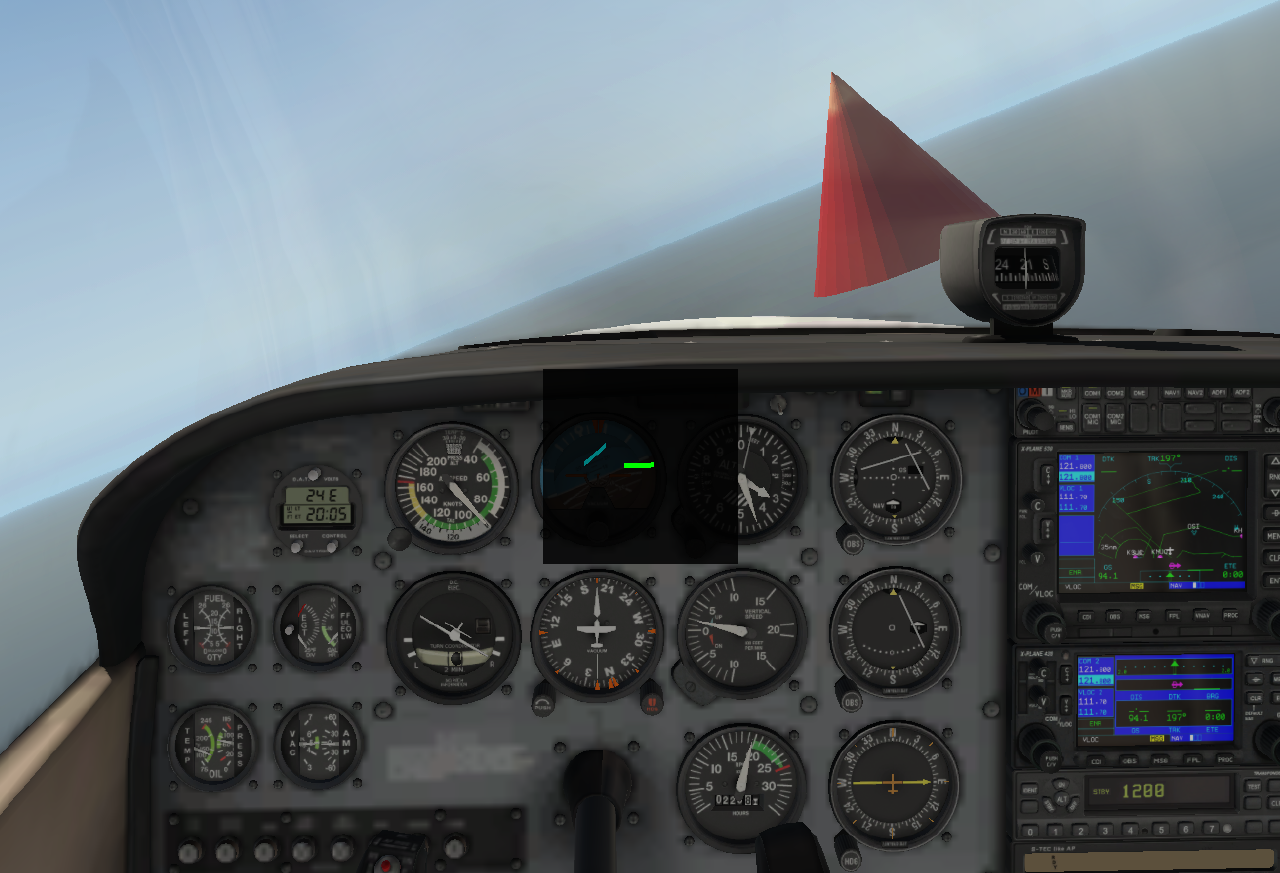} 
		\caption{Green and blue bars show visual tutor feedback}
		\label{fig:feedback}
	\end{minipage}
\end{figure}
Every time the user exceeds a threshold, the system displays a black square containing two lines denoting the user and the agent's status (Figure~\ref{fig:feedback}). The position $(x,y)$ of the middle of each line corresponds to $(r_a(t), p_a(t))$ and $(r_s(t), p_s(t))$ respectively, and the slope represents the roll angle.

This visualization shows how far apart the correct trajectory and position are. Instinctively, the student would aim to overlap both lines. Confirming the effectiveness of this and other types of feedback is part of future work.
 
\section{Conclusion and Future Work}
We presented an intelligent tutoring framework to autonomously train pilots inside a flight simulator using a simulated teacher. The teacher can provide different kinds of visual feedback to help students correct their mistakes. As future work, we will extend the simulated teacher to learn other complicated flight maneuvers like climbing and turning. We also plan to replace the BC teacher with reinforcement learning so that it can discover new policies not directly mentioned by the pilots. Lastly, we will study the impact of various kinds of feedback on student learning.

\bibliography{aaai23.bib}

\begin{thebibliography}{12}
\providecommand{\natexlab}[1]{#1}

\bibitem[{Bignold et~al.(2021)Bignold, Cruz, Taylor, Brys, Dazeley, Vamplew,
  and Foale}]{bignold21}
Bignold, A.; Cruz, F.; Taylor, M.~E.; Brys, T.; Dazeley, R.; Vamplew, P.; and
  Foale, C. 2021.
\newblock A conceptual framework for externally-influenced agents: an assisted
  reinforcement learning review.
\newblock \emph{Journal of Ambient Intelligence and Humanized Computing}.

\bibitem[{Christiano et~al.(2017)Christiano, Leike, Brown, Martic, Legg, and
  Amodei}]{ChristianoLBMLA17}
Christiano, P.~F.; Leike, J.; Brown, T.~B.; Martic, M.; Legg, S.; and Amodei,
  D. 2017.
\newblock Deep Reinforcement Learning from Human Preferences.
\newblock In \emph{Neural Information Processing Systems}, 4299--4307.

\bibitem[{Cui et~al.(2021)Cui, Koppol, Admoni, Niekum, Simmons, Steinfeld, and
  Fitzgerald}]{cui:21}
Cui, Y.; Koppol, P.; Admoni, H.; Niekum, S.; Simmons, R.~G.; Steinfeld, A.; and
  Fitzgerald, T. 2021.
\newblock Understanding the Relationship between Interactions and Outcomes in
  Human-in-the-Loop Machine Learning.
\newblock In \emph{IJCAI}, 4382--4391.

\bibitem[{Da~Silva\textbf{*} et~al.(2020)Da~Silva\textbf{*}, Hernandez-Leal,
  Kartal, and Taylor}]{da2020uncertainty}
Da~Silva\textbf{*}, F.~L.; Hernandez-Leal, P.; Kartal, B.; and Taylor, M.~E.
  2020.
\newblock Uncertainty-Aware Action Advising for Deep Reinforcement Learning
  Agents.
\newblock In \emph{Proceedings of AAAI Conference on Artificial Intelligence}.

\bibitem[{Georgila et~al.(2019)Georgila, Core, Nye, Karumbaiah, Auerbach, and
  Ram}]{georgila2019using}
Georgila, K.; Core, M.~G.; Nye, B.~D.; Karumbaiah, S.; Auerbach, D.; and Ram,
  M. 2019.
\newblock Using reinforcement learning to optimize the policies of an
  intelligent tutoring system for interpersonal skills training.
\newblock In \emph{Proceedings of the 18th International Conference on
  Autonomous Agents and MultiAgent Systems}.

\bibitem[{Morales and Sammut(2004)}]{morales2004learning}
Morales, E.~F.; and Sammut, C. 2004.
\newblock Learning to fly by combining reinforcement learning with behavioural
  cloning.
\newblock In \emph{Proceedings of the twenty-first international conference on
  Machine learning}, 76.

\bibitem[{{Oliver Wyman}(2022)}]{Wyman:22}
{Oliver Wyman}. 2022.
\newblock After COVID-19, Aviation Faces a Pilot Shortage.
\newblock
  \url{https://www.oliverwyman.com/our-expertise/insights/2021/mar/after-covid-19-aviation-faces-a-pilot-shortage.html}.
\newblock Accessed: 2022-09-15.

\bibitem[{Pomerleau(1988)}]{pomerleau1988alvinn}
Pomerleau, D.~A. 1988.
\newblock Alvinn: An autonomous land vehicle in a neural network.
\newblock \emph{Advances in neural information processing systems}, 1.

\bibitem[{Raffin et~al.(2019)Raffin, Hill, Ernestus, Gleave, Kanervisto, and
  Dormann}]{stable-baselines3}
Raffin, A.; Hill, A.; Ernestus, M.; Gleave, A.; Kanervisto, A.; and Dormann, N.
  2019.
\newblock Stable Baselines3.
\newblock \url{https://github.com/DLR-RM/stable-baselines3}.

\bibitem[{Sandstr{\"o}m, Luotsinen, and Oskarsson(2022)}]{sandstrom2022fighter}
Sandstr{\"o}m, V.; Luotsinen, L.; and Oskarsson, D. 2022.
\newblock Fighter Pilot Behavior Cloning.
\newblock In \emph{2022 International Conference on Unmanned Aircraft Systems
  (ICUAS)}, 686--695. IEEE.

\bibitem[{Shute(2008)}]{shute:08}
Shute, V.~J. 2008.
\newblock Focus on formative feedback.
\newblock \emph{Review of educational research}, 78(1): 153--189.

\bibitem[{Wang(2018)}]{wang2018reinforcement}
Wang, F. 2018.
\newblock Reinforcement learning in a pomdp based intelligent tutoring system
  for optimizing teaching strategies.
\newblock \emph{International Journal of Information and Education Technology}.

\end{thebibliography}
\end{document}